\def\year{2018}\relax
\documentclass[letterpaper]{article} 
\usepackage{aaai18}  
\usepackage{times}  
\usepackage{helvet}  
\usepackage{courier}  
\usepackage{url}  
\usepackage{graphicx}  
\usepackage{subfigure}
\usepackage{multicol}
\usepackage{multirow}
\usepackage{amsmath,amssymb}

\title{Fine Grained Knowledge Transfer for \\Personalized Task-oriented Dialogue Systems}

\author{Kaixiang Mo$^\dagger$ \and Yu Zhang$^\dagger$ \and Qiang Yang$^\dagger$ \and Pascale Fung$^\ddagger$ \\
         Hong Kong University of Science and Technology, Hong Kong, China \\ 
         {$^\dagger$\{kxmo, zhangyu, qyang\}@cse.ust.hk}, $^\ddagger$pascale@ece.ust.hk }

\date{}

\begin{document}

\maketitle

\begin{abstract}
Training a personalized dialogue system requires a lot of data, and the data collected for a single user is usually insufficient. One common practice for this problem is to share training dialogues between different users and train multiple sequence-to-sequence dialogue models together with transfer learning.
However, current sequence-to-sequence transfer learning models operate on the entire sentence, which might cause negative transfer if different personal information from different users is mixed up.
We propose a personalized decoder model to transfer finer granularity phrase-level knowledge between different users while keeping personal preferences of each user intact. A novel personal control gate is introduced, enabling the personalized decoder to switch between generating personalized phrases and shared phrases. The proposed personalized decoder model can be easily combined with various deep models and can be trained with reinforcement learning.
Real-world experimental results demonstrate that the phrase-level personalized decoder improves the BLEU over multiple sentence-level transfer baseline models by as much as $7.5\%$.
\end{abstract}

\section{Introduction}
\label{Intro}
\noindent Task-oriented dialogue systems aim to help a user to finish a certain task with dialogues, and it can be categorized into rule-based systems and learning-based systems. Learning-based dialogue systems do not require the predefined dialogue state, which are more general and suitable for the situation when exact dialogue states are hard to define. In particular, neural network based dialogue systems do not require the predefined templates and are more flexible. In this paper, we focus on neural network based task-oriented dialogue systems. 

Personalized dialogue systems can help the target user to complete a task faster and make the dialogue more efficient. In a personalized dialogue system, the personal information and preference of the target user are recorded and utilized, thus the personal dialogue policy can generate personalized sentences and speed up the dialogue process for the target user. 
In personalized dialogue systems, the dialogue policy and response generation for each user is different. Training a whole personalized dialogue system requires a lot of data and the data collected from a single user is usually insufficient. Multi-tasking can be used to transfer dialogue knowledge across different users by sharing training dialogues. 

However, current neural network transfer models work on the entire sentence, but the phrases concerning personal information from different users should not be transferred across users at all. For example, if the dialogue data from a sugar lover is used to train a personalized dialogue system for a diabetes patient, the system might recommend sweet drinks to the patient and cause a disaster. Hence, we think that knowledge transfer in personalized dialogue systems should be conducted in finer granularity.

\begin{figure}[t]
\centering\mbox{
\scalebox{1.0}{\includegraphics[width=\columnwidth]{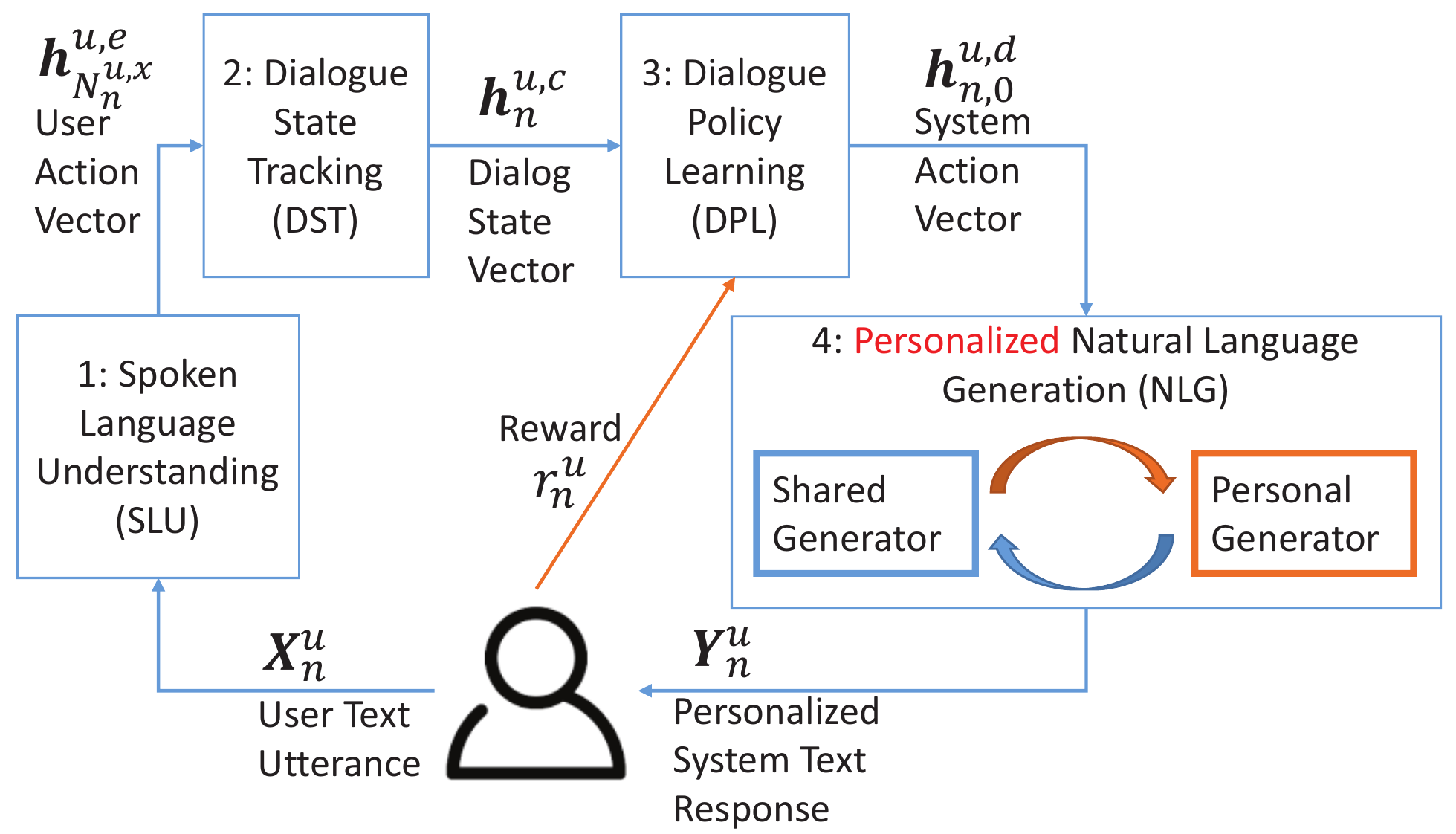}}
}
\caption{Framework for Personalized Task-oriented Dialogue System}
\label{fig:DialogFramework}
\end{figure}

In this paper, we propose a personalized decoder that can transfer shared phrase-level knowledge between \mbox{different} users while keeping the personalized information of each user intact.
A novel personal control gate is introduced in the proposed personalized decoder, enabling the decoder to switch between generating shared phrases and personal phrases.
For example in Figure~\ref{fig:decoder_personal}, ``Hot Latte'' is a personal phrase and ``Still'' and ``?'' are shared phrases. The personalized decoder can generate different personal phrases for different users in a sentence, while the knowledge for the shared phrases are shared among all users.
The proposed personalized decoder can be easily used with various base models to achieve fine-grained phrase-level dialogue knowledge transfer between different users. Real-world experimental results show that the personalized decoder can improve the BLEU score over multiple sentence-level transfer models by as much as $7.5\%$.


\section{Related work}
\label{sect:related}
Learning based task-oriented dialogue systems~\cite{casanueva2015knowledge,genevay2016transfer,gavsic2013pomdp,gavsic2014incremental,williams2016end,bordes2016learning} can select appropriate answers from a set of predefined answers.
Ga\v{s}ic et al.~\cite{gavsic2013pomdp,gavsic2014incremental} used transfer learning in the dialogue \mbox{management} module to extend a dialogue system to handle previous unseen concepts. Williams et al.~\cite{williams2016end,bordes2016learning} used learning methods in spoken language understanding and dialogue management. These methods require the predefined answer candidates or templates.
Wen~\cite{wen2015semantically,wen2016conditional,wen2016network} proposed to use trainable modules for each part of the dialogue system which does not require predefined answer candidates or templates. However, this model still requires the predefined slots for dialogue policy learning, which limits its application.

Neural based dialogue response generation systems require neither the answer candidates/templates, nor the predefined slots for dialogue policy. Sequence to sequence (seq2seq) models and their variants~\cite{bahdanau2014neural,sutskever2014sequence,sordoni2015neural,serban2015building,li2016deep} are widely used to model dialogues. However, data collected from each user is insufficient for training a personalized dialogue system.

Multi-task learning is used to transfer knowledge in sequence to sequence model. Luong et al.~\cite{luong2015multi} proposed to share encoder/decoder across different tasks in order to achieve knowledge transfer. However, these knowledge transfer works on the granularity of sentences. Since different persons have different preferences and require different dialogue policies and responses, directly transferring sentences across different users might lead to the negative transfer.

In this paper, a personalized decoder is proposed for response generation, which is capable of transferring dialogue knowledge, and it can easily be combined with many models including seq2seq~\cite{sutskever2014sequence} and HRED~\cite{serban2015building}.

\section{Problem}
\label{sect:Problem}
In this section, we first define notations and then present the problem settings.

\subsection{Notation}
In this paper, matrices are denoted in a bold capital case, column vectors are in a bold lower case, and scalars are in a lower case.
Questions are denoted by $X=\{ \mathbf{X}^u_n \}_{n=1}^N$, where $N$ is the number of turns in a dialogue, the superscript $u$ denotes users, $\mathbf{X}^u_n=\{{x}^u_{n,t}\}_{t=1}^{{N^{u,x}_n}}$ denotes the $n$-th question, and $N^{u,x}_n$ is the number of words in $\mathbf{X}^u_n$. Responses are denoted by $Y=\{ \mathbf{Y}^u_n \}_{n=1}^N$ and the $n$-th response $\mathbf{Y}^u_n$ is denoted by $\mathbf{Y}^u_n=\{ {y}^u_{n,t} \}_{t=1}^{{N^{u,y}_n}}$, where $N^{u,y}_n$ is the number of words in $\mathbf{Y}^u_n$. To be consistent, a dialogue turn is indexed by $n$ and a word is indexed by $t$.

\subsection{Problem Definition}
Given the conversation history of multiple users, we aim to learn an end-to-end personalized task-oriented dialogue system for each user.
The input of the problem are:
\begin{enumerate}
\item Historical dialogue sessions ${\mathcal{T}^u}=\{ \mathbf{X}_n^u, \mathbf{Y}_n^u, r_n^u \}_{n=1}$ of each user, where $r_n^u$ is the reward obtained at $n$-th dialogue turn.
\item Personal word label $\mathcal{O}_n^u=\{ o_{n,t}^u \}_{t=1}^N$ for each word in $\mathbf{Y}_n^u$, where $o_{n,t}^u=1$ means that $x_{n,t}^u$ is a personal word and $o_{n,t}^u=0$ means the word $x_{n,t}^u$ is a general word.
\end{enumerate}
The output of the problem are:
\begin{enumerate}
\item A dialogue policy $\pi^u$ for each user $u$, which generates a response $\mathbf{Y}_n^u$ for each dialogue history $\mathcal{H}^u_n=\{ \{ \mathbf{X}_i^u, \mathbf{Y}_i^u \}_{i=1}^{n-1}, \mathbf{X}_n^u \}$.
\end{enumerate}

\begin{figure}[t]
\centering\mbox{
\subfigure[Response generation with a basic decoder. $\mathbf{h}^{u,c}_{n}$ is not shown since it is the same for all time step $t$.]{\scalebox{0.9}{\includegraphics[width=\columnwidth]{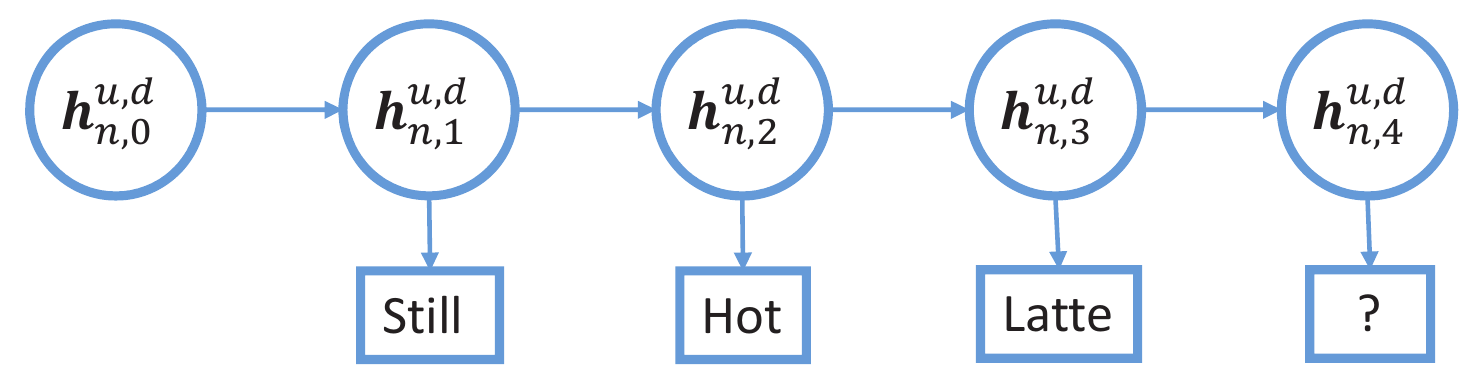}} \label{fig:decoder_normal} }

}
\centering\mbox{
\subfigure[Personalized response generation with the proposed personalized decoder. The personal control gate at different time steps are denoted in gray circles.]{\scalebox{0.9}{\includegraphics[width=\columnwidth]{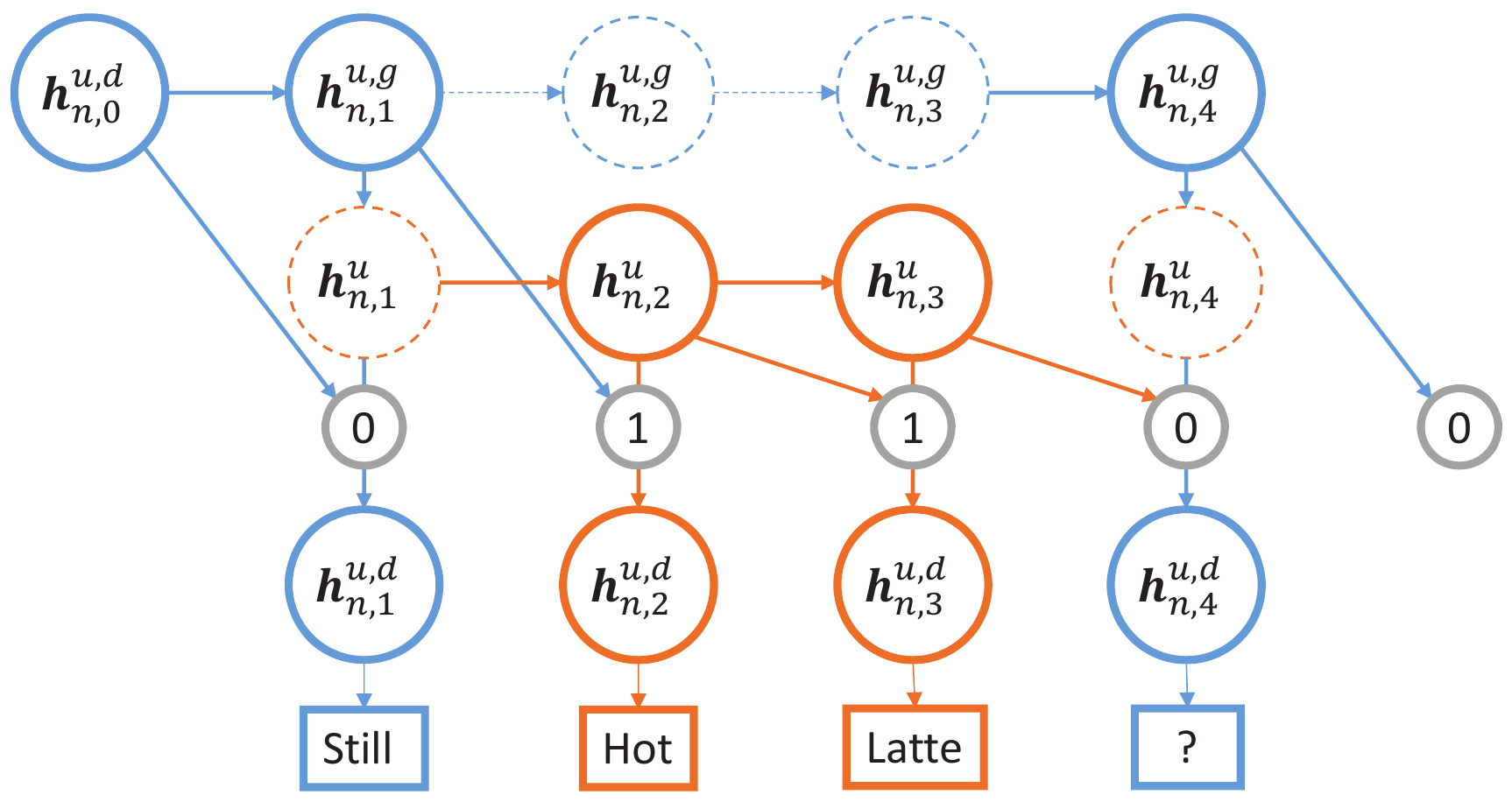}} \label{fig:decoder_personal} }

}
\caption{Response generation comparison}
\label{fig:decoder}
\end{figure}

\section{Personalized Decoder}
\label{sect:Model}
In this section, we firstly introduce the proposed personalized decoder, then introduce the combination of the personalized decoder with two models, and finally present the training method.

\subsection{Decoder for Response Generation}

In this section, we introduce the mathematical formulation of the proposed personalized decoder.

\subsubsection{Basic Decoder}

The hidden state for the $t$-th word in the $n$-th turn is defined as
$$ \mathbf{h}^{u,d}_{n,t} = \tanh(\mathbf{W}^d \mathbf{h}^{u,d}_{n,t-1} + \mathbf{U}^d \mathbf{e}_{\hat{y}^u_{n,t-1}} + \mathbf{V}^d \mathbf{h}^{u,c}_{n}),$$
where $\mathbf{e}_{\hat{y}^u_{n,t-1}}$ is the word embedding of the last word $\hat{y}^u_{n,t-1}$ in the same sentence, and $\tanh(\cdot)$ denotes the hyperbolic tangent function. The decoder RNN in Figure~\ref{fig:decoder_normal} takes $\mathbf{h}^{u,d}_{n,0}$ and $\mathbf{h}^{u,c}_{n}$ as inputs\footnote{$\mathbf{h}^{u,c}_{n}$ is not shown in Figure~\ref{fig:decoder_normal} since it is the same for all time step $t$.} and then generates the response word by word, where $\mathbf{h}^{u,c}_{n}$ is the embedding vector used to generate the current response. The probability of generating the next word $\hat{y}^u_{n,t}=y^u_{n,t}$ given $\mathbf{h}^{u,d}_{n,t}$ and $\hat{y}^u_{n,t-1}$ is
{\small
\begin{align}
\label{eq:1}
&\omega(\mathbf{h}^{u,d}_{n,t}, \hat{y}^u_{n,t-1}) = \mathbf{H}_0 \mathbf{h}^{u,d}_{n,t} + \mathbf{E}_0 \mathbf{e}_{\hat{y}^u_{n,t-1}} + \mathbf{b}_o \\
\label{eq:2}
&g(\mathbf{h}^{u,d}_{n,t}, \hat{y}^u_{n,t-1}, y^u_{n,t}) = \mathbf{o}_{y^u_{n,t}}^T \omega(\mathbf{h}^{u,d}_{n,t}, \hat{y}^u_{n,t-1}) \\
\label{eq:3}
&p(\hat{y}^u_{n,t}=y^u_{n,t}) = \frac{\exp(g(\mathbf{h}^{u,d}_{n,t}, \hat{y}^u_{n,t-1}, y^u_{n,t}))}{\sum_{\forall y'} \exp(g(\mathbf{h}^{u,d}_{n,t}, \hat{y}^u_{n,t-1}, y')) }
\end{align}
}\noindent
where $\mathbf{o}_{v}$ is the output embedding for word $v$, and $\mathbf{H}_0$, $\mathbf{E}_0$ and $\mathbf{b}_o$ are parameters.

\subsubsection{Personalized Dialogue Decoder for Phrase-level Transfer Learning}
\label{subsect:personal_decoder}
In this section, we present the proposed personalized decoder as illustrated in Figure~\ref{fig:decoder_personal}.
While the sentence-level transfer is to transfer entire sentences, the proposed personalized decoder is on the phrase level and is to transfer a shared fraction of the sentences to the target domain, where a phrase is a short sequence of words containing a coherent meaning, for example, an address.
In order to achieve maximum knowledge transfer and to avoid negative transfer caused by differences in user preferences, the proposed personalized decoder has a shared component and a personalized component. In order to learn to switch between the shared and personal components in the phrase level, we introduce a personal control gate $o^u_{n,t}$ for each word, which is learned from the training data.

Given the the embedding vector for the $n$-th response $\mathbf{h}^{u,c}_{n}$ and initial hidden state $\mathbf{h}^{u,d}_{n,0}$ for word $\hat{y}^u_{n,0}$, the initial states are computed as
\begin{align*}
&\mathbf{h}^{u,g}_{n,0} = \mathbf{h}^{u,d}_{n,0}, \mathbf{h}^u_{n,0} = \mathbf{h}^{u,d}_{n,0}, \hat{o}^u_{n,0}=0\\
&\mathbf{e}_{\hat{y}^u_{n,0}} = 0, \mathbf{e}_{\hat{y}^{u,g}_{n,0}} = 0
\end{align*}
where $\mathbf{h}^{u,g}_{n,t}$ is the hidden state for the shared component, and $\hat{y}^{u,g}_{n,t}$ records the last word generated by the shared component, $\mathbf{h}^u_{n,t}$ is the hidden state for the personal component, and $\mathbf{h}^{u,d}_{n,t}$ is the hidden state for generating the word $\hat{y}^u_{n,t}$.

The shared component adopts the GRU model to capture the long-term dependency and is shared by all users. Specifically, at each time step $t$, the shared component is defined as
{\small
\begin{eqnarray*}
&\mathbf{z}^u_{n,t} = \sigma(\mathbf{W}^g_z \mathbf{h}^{u,g}_{n,t-1} + \mathbf{U}^g_z \mathbf{e}_{\hat{y}^{u,g}_{n,t-1}} + \mathbf{V}^g_z \mathbf{h}^{u,c}_n + \mathbf{b}_z) \label{eq:gen_z}\\
&\mathbf{r}^u_{n,t} = \sigma(\mathbf{W}^g_r \mathbf{h}^{u,g}_{n,t-1} + \mathbf{U}^g_r \mathbf{e}_{\hat{y}^{u,g}_{n,t-1}} + \mathbf{V}^g_r \mathbf{h}^{u,c}_n + \mathbf{b}_r) \label{eq:gen_r}\\
&\mathbf{\tilde{h}}^{u,g}_{n,t} = \sigma(\mathbf{W}^g_h (\mathbf{r}^u_{n,t} \odot \mathbf{h}^{u,g}_{n,t-1}) + \mathbf{U}^g_h \mathbf{e}_{\hat{y}^{u,g}_{n,t-1}} + \mathbf{V}^g_h \mathbf{h}^{u,c}_n + \mathbf{b}_{h} ) \label{eq:gen_h_tild}\\
&\mathbf{\hat{h}}^{u,g}_{n,t} = \mathbf{z}^u_{n,t} \odot \mathbf{h}^{u,g}_{n,t-1} + (1-\mathbf{z}^u_{n,t}) \odot \mathbf{\tilde{h}}^{u,g}_{n,t}, \label{eq:gen_h_hat}
\end{eqnarray*}
}\noindent
where $\odot$ denotes the element-wise product between vectors or matrices, $\sigma(\cdot)$ is the sigmoid function, $\mathbf{z}^u_{n,t}$ is the update gate, $\mathbf{r}^u_{n,t}$ is the forget gate, and $\mathbf{\hat{h}}^{u,g}_{n,t}$ is the tentative updated hidden state.
If the $t$-th word is a shared word (i.e., $\hat{o}^u_{n,t}=0$), then we update the shared hidden state and last general word as usual and otherwise $\mathbf{h}^{u,g}_{n,t}$ and $\mathbf{e}_{\hat{y}^{u,g}_t}$ remain unchanged. Thus $\mathbf{h}^{u,g}_{n,t}$ and $\mathbf{e}_{\hat{y}^{u,g}_t}$ can be updated as
\begin{align*}
&\mathbf{h}^{u,g}_{n,t} = (1-\hat{o}^u_{n,t})\odot\mathbf{\hat{h}}^{u,g}_{n,t} + \hat{o}^u_{n,t}\odot\mathbf{h}^{u,g}_{n,t-1}\\
&\mathbf{e}_{\hat{y}^{u,g}_t} = (1-\hat{o}^u_{n,t})\odot\mathbf{e}_{\hat{y}^u_{t-1}} + \hat{o}^u_{n,t}\odot\mathbf{e}_{\hat{y}^{u,g}_{t-1}}.
\end{align*}

The personal component is a RNN model, which generates personalized sequence based on sentence context $\mathbf{h}^{u,g}_{n,t}$ from the shared component. There is an separate RNN model for each user. At each time step $t$, the personal component receives $\mathbf{e}_{\hat{y}^u_{t-1}}$, $\hat{o}^u_{n,t}$, $\mathbf{h}^u_{n,t-1}$ and $\mathbf{h}^{u,g}_{n,t-1}$ as inputs and outputs $\mathbf{\hat{h}}^u_{n,t}$, which is defined as
$$ \mathbf{\hat{h}}^u_{n,t} = \sigma(\mathbf{W}^{u} \mathbf{h}^u_{n,t-1} + \mathbf{U}^{u} \mathbf{e}_{\hat{y}^u_{n,t-1}} + \mathbf{V}^{u} \mathbf{h}^{u,g}_{n,t-1} ).$$
The personal hidden state will be update as
$$ \mathbf{h}^u_{n,t} = (1-\hat{o}^u_{n,t})\odot\mathbf{h}^{u,g}_{n,t} + \hat{o}^u_{n,t}\odot\mathbf{\hat{h}}^u_{n,t}.$$
$\mathbf{h}^u_{n,t}$ equals $\mathbf{\hat{h}}^u_{n,t}$ if the control gate is on corresponding to $\hat{o}^u_{n,t}=1$. If $\hat{o}^u_{n,t}$ equals $0$, $\mathbf{h}^u_{n,t}$ will take the value of $\mathbf{h}^{u,g}_{n,t}$.

The personal control gate $o^u_{n,t}$ is binary, i.e., $o^u_{n,t} \in \{0,1\}$. The predicted control gate $\hat{o}^u_{n,t}$ at time $t$ is a function of $\hat{o}^u_{n,t-1}$, $\mathbf{h}^{u,g}_{n,t-1}$, $\mathbf{h}^u_{n,t-1}$, and $\mathbf{e}_{\hat{y}^u_{n,t-1}}$ as\footnote{In training process, the ground truth $o^u_{n,t}$ is used as label to train the prediction function for $\hat{o}^u_{n,t}$.}
\begin{tiny}
\begin{equation}
p(\hat{o}^u_{n,t}=1) =
\left\{
\begin{array}{cl}
\sigma(\mathbf{W}^g_o \mathbf{h}^{u,g}_{n,t-1} + \mathbf{U}^g_o \mathbf{e}_{\hat{y}^u_{n,t-1}} + \mathbf{b}_o) & \text{if } \hat{o}^u_{n,t-1}=0 \\
\sigma(\mathbf{W}^{u}_o \mathbf{h}^u_{n,t-1} + \mathbf{U}^{u}_o \mathbf{e}_{\hat{y}^u_{n,t-1}} + \mathbf{b}^u_o) & \text{if } \hat{o}^u_{n,t-1}=1
\end{array}
\right..
\label{eq:gate_o}
\end{equation}
\end{tiny}\noindent
$\hat{o}^u_{n,t}$ decides whether to use the personal component to generate the next word. $\mathbf{h}^{u,d}_{n,t}$ is defined as
\begin{equation}
\mathbf{h}^{u,d}_{n,t} = (1-\hat{o}^u_{n,t})\odot\mathbf{h}^{u,g}_{n,t} + \hat{o}^u_{n,t}\odot\mathbf{h}^u_{n,t},
\label{eq:gen_combine}
\end{equation}
where $\mathbf{h}^{u,d}_{n,t}$ is the hidden vector that directly generates the next word $\hat{y}^u_{n,t}$ and the probability of generating the next word $y^u_{n,t}$ is defined by the generation process in Eqs.~(\ref{eq:1}-\ref{eq:3}).

\begin{figure*}[t]
\centering\mbox{
\scalebox{1.3}{\includegraphics[width=\columnwidth]{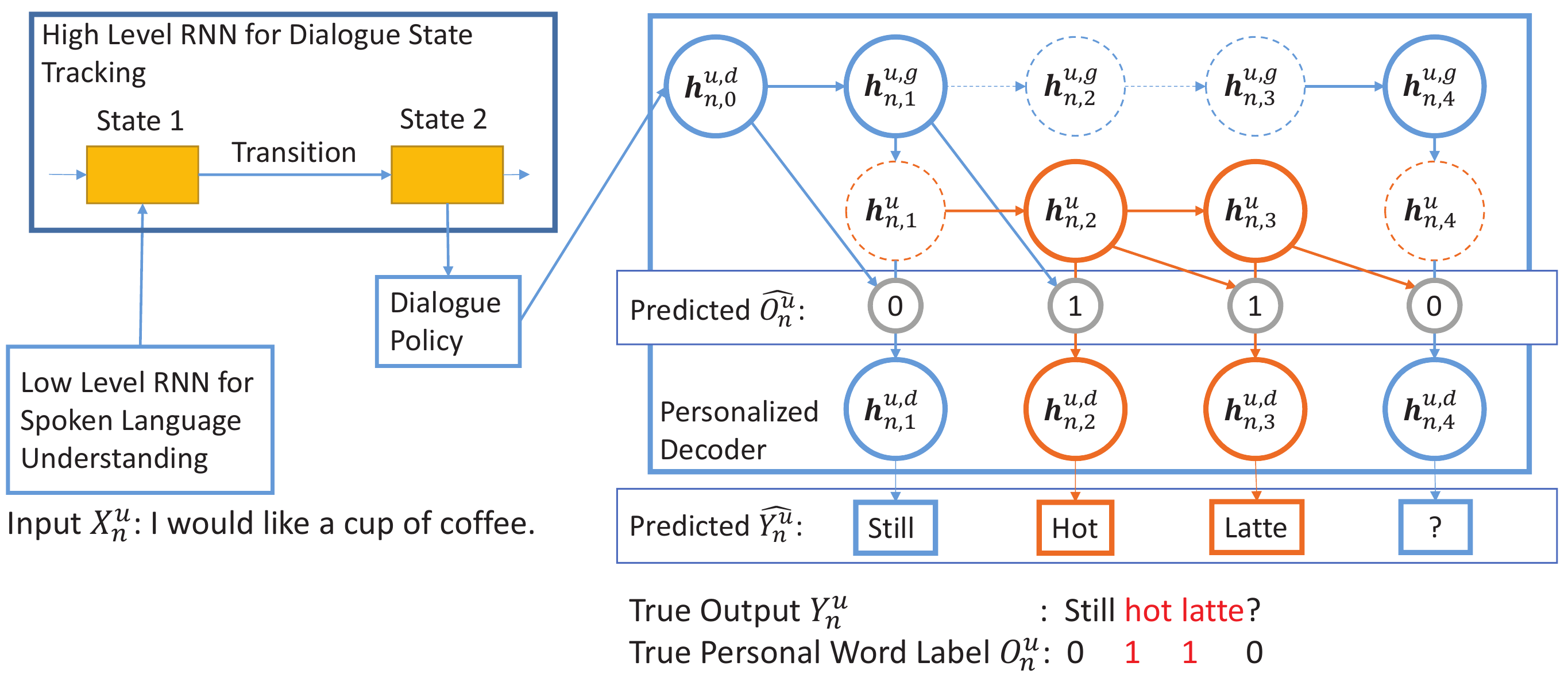}}
}
\caption{Model framework for personalized HRED, obtained by combining HRED with Personalized Decoder}
\label{fig:framework}
\vspace{-1em}
\end{figure*}

The decoding procedure is as follows:
\begin{enumerate}
\item Initialize $\mathbf{h}^{u,g}_{n,0}$, $\mathbf{h}^u_{n,0}$, $\hat{o}^u_{n,0}, \mathbf{e}_{\hat{y}^u_{n,0}}, \mathbf{e}_{\hat{y}^{u,g}_{n,0}}$ based on $\mathbf{h}^{u,d}_{n,0}$ and $\mathbf{h}^{u,c}_{n}$. $\hat{o}^u_{n,0}$ is initialized to be $0$ and $\mathbf{e}_{\hat{y}^u_{n,0}}$ is initialized to be a zero vector.
\item Compute $\hat{o}^u_{n,t}$ based on $\mathbf{h}^{u,c}_n$, $\hat{o}^u_{n,t-1}$, $\mathbf{h}^{u,g}_{n,t-1}$, $\mathbf{h}^u_{n,t-1}$ and $\mathbf{e}_{\hat{y}^u_{t-1}}$ with Eqs. (\ref{eq:gate_o}).
\item Compute $\mathbf{h}^{u,g}_{n,t}$, $\mathbf{h}^u_{n,t}$ and $\mathbf{h}^{u,d}_{n,t}$ based on $\hat{o}^u_{n,t}$. 
\item Generate $\hat{y}^u_{n,t}$ based on $\mathbf{h}^{u,d}_{n,t}$ with Eqs. (\ref{eq:gen_combine}).
\item Repeat step 2 to step 4 until the ending symbol.
\end{enumerate}
The shared and personal components can be trained together with reinforcement learning as illustrated in the parameter learning section.

Compared with the basic decoder, the personalized decoder is novel in the following aspects:
\begin{enumerate}
\item Knowledge is transferred in the phrase level rather than the sentence level.
\item There is a shared component for knowledge transfer and a personalized component for each user in order to avoid negative transfer caused by different data distributions.
\item The personal control gate can learn when to use the shared component and when to use the personal component to generate a fluent and personalized response.
\end{enumerate}

\subsection{Combining HRED with Personalized Decoder}
\label{sect:PHRED}
In this section, we show how the proposed personalized decoder can be combined with hierarchical recurrent encoder-decoder (HRED)~\cite{serban2015building} to make a personalized HRED, whose model is shown in Fig.~\ref{fig:framework}.

In the personalized HRED, there are a low-level encoding RNN, a high-level RNN, a dialogue policy and a personalized decoder. The low-level encoding RNN for spoken language understanding encodes user utterances into a user action vector, the high-level RNN is responsible for dialogue state tracking, the dialogue policy maps the dialogue state to the action vector, and the personalized decoder is responsible for generating words as the response.

\subsubsection{Spoken Language Understanding with Word Encoder}
The low-level encoder RNN for spoken language understanding maps each $\mathbf{X}^u_n=\{{x}^u_{n,1}, {x}^u_{n,2}, \cdots, {x}^u_{n,N^{u,x}_n} \}$ to a fixed dimension vector $\mathbf{h}^{u,e}_{N^{u,x}_n}$ as
$$ \mathbf{h}^{u,e}_{n,t} = \tanh(\mathbf{W}^e \mathbf{h}^{u,e}_{n,t-1} + \mathbf{U}^e \mathbf{e}_{x^u_{n,t}}), $$
where $\mathbf{U}^e$ is the input embedding matrix and $\mathbf{W}^e$ is the weight matrix corresponding to hidden state (word-level) transition function.
Similarly we can define a mapping from the response $\mathbf{Y}^u_n=\{{y}^u_{n,1}, {y}^u_{n,2}, \cdots, {y}^u_{n,N^{u,y}_n} \}$ to the response vector $\bar{\mathbf{h}}^{u,e}_{N^{u,x}_n}$ as
$$ \bar{\mathbf{h}}^{u,e}_{n,t} = \tanh(\mathbf{W}^e \bar{\mathbf{h}}^{u,e}_{n,t-1} + \mathbf{U}^e \mathbf{e}_{y^u_{n,t}}). $$

\subsubsection{Dialogue State Tracking with Sentence RNN}
The high-level sentence RNN tracks dialogue states based on all previous sentences in the dialogue. This RNN takes $\bar{\mathbf{h}}^{u,e}_{n-1, N^{u,y}_{n-1}}$, $\mathbf{h}^{u,e}_{n, N^{u,x}_{n}}$ and ${\mathbf{h}}^{u,c}_{n-1}$ as the input to generate next dialogue state $\mathbf{h}^{u,c}_{n}$ as:
\begin{align*}
&\bar{\mathbf{h}}^{u,c}_{n-1} = \tanh(\mathbf{W}^c \mathbf{h}^{u,c}_{n-1} + \mathbf{U}^c \bar{\mathbf{h}}^{u,e}_{n-1,N^{u,y}_{n-1}})\\
&\mathbf{h}^{u,c}_{n} = \tanh(\mathbf{W}^c \bar{\mathbf{h}}^{u,c}_{n} + \mathbf{U}^c \mathbf{h}^{u,e}_{n,N^{u,x}_n}),
\end{align*}
where $\bar{\mathbf{h}}^{u,e}_{n-1,N^{u,y}_{n-1}}$ and $\mathbf{h}^{u,e}_{n,N^{u,x}_n}$ are encoding vectors of sentences $\mathbf{Y}^u_{n-1}$ and $\mathbf{X}^u_n$, $\mathbf{U}^c$ is the embedding matrix for the input, and $\mathbf{W}^c$ is the weight matrix corresponding to the hidden state (sentence-level) transition function.

\subsubsection{Dialogue Policy with Linear Transformation}
Given the current dialogue state $\mathbf{h}^{u,c}_{n}$, the action vector $\mathbf{h}^{u,d}_{n,0}$ is calculated as
$$ \mathbf{h}^{u,d}_{n,0} = \tanh(\mathbf{D}_0 \mathbf{h}^{u,c}_{n} + \mathbf{b}_0) $$
where $\mathbf{D}_0$ and $\mathbf{b}_0$ are the policy parameters.

\subsubsection{Response Generation with Personalized Decoder}
Then the personalized decoder takes $\mathbf{h}^{u,c}_{n}$ and $\mathbf{h}^{u,d}_{n,0}$ as inputs to generate the response word by word.

The joint probability of $(\mathcal{H}^u_n, \mathcal{O}^u_n, \mathbf{Y}^u_n)$ is given by:
\begin{align*}
&p(\mathcal{H}^u_n, \mathcal{O}^u_n, \mathbf{Y}^u_n) = p(\mathcal{H}^u_n)p(\mathcal{O}^u_n, \mathbf{Y}^u_n | \mathcal{H}^u_n)\\
&p(\mathcal{H}^u_n) = \prod_k^{n-1} p(\mathcal{O}^u_k, \mathbf{Y}^u_k | \mathcal{H}^u_k),
\end{align*}
where $\mathcal{O}^u_n=\{ o^u_{n,t} \}_{t=0}^{N^{u,y}_n}$ is the collection of control gate variables for sentence $\mathbf{Y}^u_n$.

\subsection{Combining Seq2Seq Model with Personalized Decoder}
The proposed personalized decoder can be combined with the popular seq2seq model proposed in \cite{sutskever2014sequence}. We can build a personalized seq2seq model by replacing the original decoder in the seq2seq model with the proposed personalized decoder. The whole encoder and the common component of the personalized decoder are shared across all users, while each user has his own personal component in the personalized decoder.

Moreover, the proposed personalized decoder can also be easily combined with many other models to achieve knowledge transfer.

\subsection{Parameter Learning}
\label{sect:Parameter_Learning}
The whole model is trained in an end-to-end manner with reinforcement learning~\cite{sutton1998reinforcement} such that the model will generate personalized system response according to current dialogue state to maximize the future cumulative reward.

\subsubsection{Reward}
\label{subsect:reward}
The agent will receive general rewards and personal rewards, and the total reward is the sum of general reward and personal reward. The general and personal rewards will be received under the following conditions: 
\begin{enumerate}
\item Personal rewards of $0.3$ will be received when the user confirms the suggestion of the agent. This is related to the personal information of the user. For example, the user could confirm the address suggested by the agent.
\item General rewards of $0.1$ will be received when the user provides the information about the target task.
\item General rewards of $1.0$ will be received when the system helps the user finish the target task successfully.
\item General reward of $-0.2$ will be received by the agent when the user rejects to proceed if the system is generating non-logical responses. A reward of $-0.05$ will be received for each the dialogue turn.
\end{enumerate}

\subsubsection{Loss function}
We use the policy gradient method to learn the parameter of the model. Specifically, we use the REINFORCE~\cite{williams1992simple} algorithm. The training set consists of a set of trajectories $\mathcal{T}^u=\{\mathcal{H}_n^u,\mathcal{O}^u_n,\mathbf{Y}_n^u,r_n^u\}$, which records actions provided by the user. We denote all these parameters by $\Theta$, and the model policy is denoted by $\pi$.

We define the return of the trajectory $J^{\mathcal{T}^u}=\sum_{n=1}^N \gamma^{n-1} r_n^u$ where $r_n^u$ is the reward obtained at $n$-th dialogue turn. Then the loss function is defined as the expected reward of the policy under all trajectory $\mathcal{T}^u$:
$$ J(\Theta) = \int_{\mathcal{T}^u} p(\mathcal{T}^u|\Theta) d \mathcal{T}^u,$$ whose gradient can be computed as
{\small
$$ \Delta_{\Theta} J(\Theta) = \mathbb{E}\{ \sum_{n=0}^N \Delta_{\Theta} \log \pi_{\Theta}(\mathcal{H}_n^u, \mathcal{O}^u_n, \mathbf{Y}_n^u)  J^{\mathcal{T}^u}_n \}$$
}\noindent
where $J^{\mathcal{T}^u}_n=\sum_{k=n}^N \gamma^{k-n} r_k^u$ is the future cumulative reward and $N$ is the total number of dialogue turns.

\subsubsection{Optimization}
We adopt the Adam algorithm~\cite{kingma2014adam} to optimize our model. We train the model on data collected from different users, where the shared parameters are updated in each iteration while the personalized parameters are updated based on data collected from the corresponding user only.


\section{Experiments}
\label{sect:Experiment}
In this section, we experimentally verify the effectiveness of the personalized decoder by conducting experiments on a real-world dataset and a simulation dataset. 

We compare the proposed two personalized phrase-level transfer models with their sentence-level counterparts and none-transfer versions. Note that we do not assume the predefined dialogue states or templates, thus rule-based systems do not apply here. All methods are listed as follows:
\begin{enumerate}
\item None-transfer seq2seq \cite{sutskever2014sequence} (denoted by ``S2S''): The sequence to sequence model is trained only with data from each individual user without transfer learning.
\item Sentence-level transfer seq2seq \cite{luong2015multi} (denoted by ``ST-S2S''): The sequence to sequence model trained in a multi-task setting with both the encoder and the decoder shared across all users. Sentence-level knowledge is transferred.
\item Sentence-level encoder transfer seq2seq \cite{luong2015multi} (denoted by ``ST-E-S2S''): The sequence to sequence model is trained in a multi-task setting with the encoder shared across all users but a different decoder for each target user. Sentence-level knowledge is transferred.
\item Personalized phrase-level transfer seq2seq (denoted by ``PT-S2S''): The sequence to sequence model is equipped with the proposed personalized decoder and trained in the multi-task setting. Phrase-level knowledge is transferred.
\item None-transfer HRED \cite{serban2015building} (denoted by ``HRED''): The HRED model is trained only with data from each individual user, without transfer learning.
\item Sentence-level transfer HRED (denoted by ``ST-HRED''): The HRED model is trained in a multi-task setting, while sentence-level knowledge is transferred.
\item Personalized phrase-level transfer HRED (denoted by ``PT-HRED''): The personalized HRED model with the proposed personalized decoder is trained in a multi-task setting. Phrase-level knowledge is transferred.
\end{enumerate}

\begin{table}[!ht]
\small
\caption{Statistics of the dataset}
\begin{center}
\begin{tabular}{|c||c|c|c|c|}
  \hline
   & \multicolumn{2}{|c|}{Train Set} & \multicolumn{2}{|c|}{Test Set} \\
  \hline
  Dataset & Users & Dialogues & Users & Dialogues \\
  \hline
  Simulation & 10 & 50 & 10 & 2,000 \\
  \hline
  Real & 52 & 464 & 52 & 831 \\
  \hline

\end{tabular}
\label{tab:DataTab}
\end{center}
\vspace{-2em}
\end{table}

\subsection{Simulation Experiments}
In this section, we conduct experiments on a simulation dataset.

We build a rule-based user simulator for multi-turn coffee ordering services. The user simulator will order his favorite coffee with probability 0.8 and try new coffee with probability 0.2. The simulated user will answer questions about the coffee type, the temperature, the cup size and the delivery address. The user will give rewards according to the reward function defined in the reward section. All models will select an answer with the highest generative probability from a set of answer candidate templates, which is filled with the confirmed information obtained from the simulation user.
In order to obtain ground truth dialogues, we design a rule-based ground truth agent which will choose the best reply with probability 0.8 and choose a random reply with probability 0.2. We have 10 simulated users in total. For each user, 5 dialogues are collected for training and 200 dialogues are collected for testing. In the training and testing processes, \mbox{all} interior ground truth information and slot information will not be used.

We use the BLEU~\cite{papineni2002bleu} and slot-error-rate as the off-line evaluation metric, and use the averaged reward and averaged success rate as the online evaluation metrics in order to fully evaluate our models. The original slot-error-rate is the ratio of the number of wrong-slots and missing-slots over the total number of slots in the given ground truth dialogue act. However, in our setting, the ground truth dialogue act is not given before response generation, thus the missing-slots do not make sense. We modify the definition of slot-error-rate so that it can adapt to our setting. The wrong-slots are defined as the slots not in the simulated user's preference.
Hence we define the slot-error-rate as the ratio of the number of wrong-slots over the total number of slots in the generated sentence. Note that a low slot-error-rate doesn't necessarily mean the responses are fluent or appropriate, it only means fewer wrong-slots in the responses. For online testing, each simulated user will generate 1000 coffee orders and the mean and standard deviation of rewards obtained in the dialogues are reported. We run the experiments with 5 different random seeds and we report the mean and standard deviation of the BLEU, slot error rate, averaged rewards and success rate.

\begin{figure}[t]
\centering\mbox{
\scalebox{1.0}{\includegraphics[width=\columnwidth]{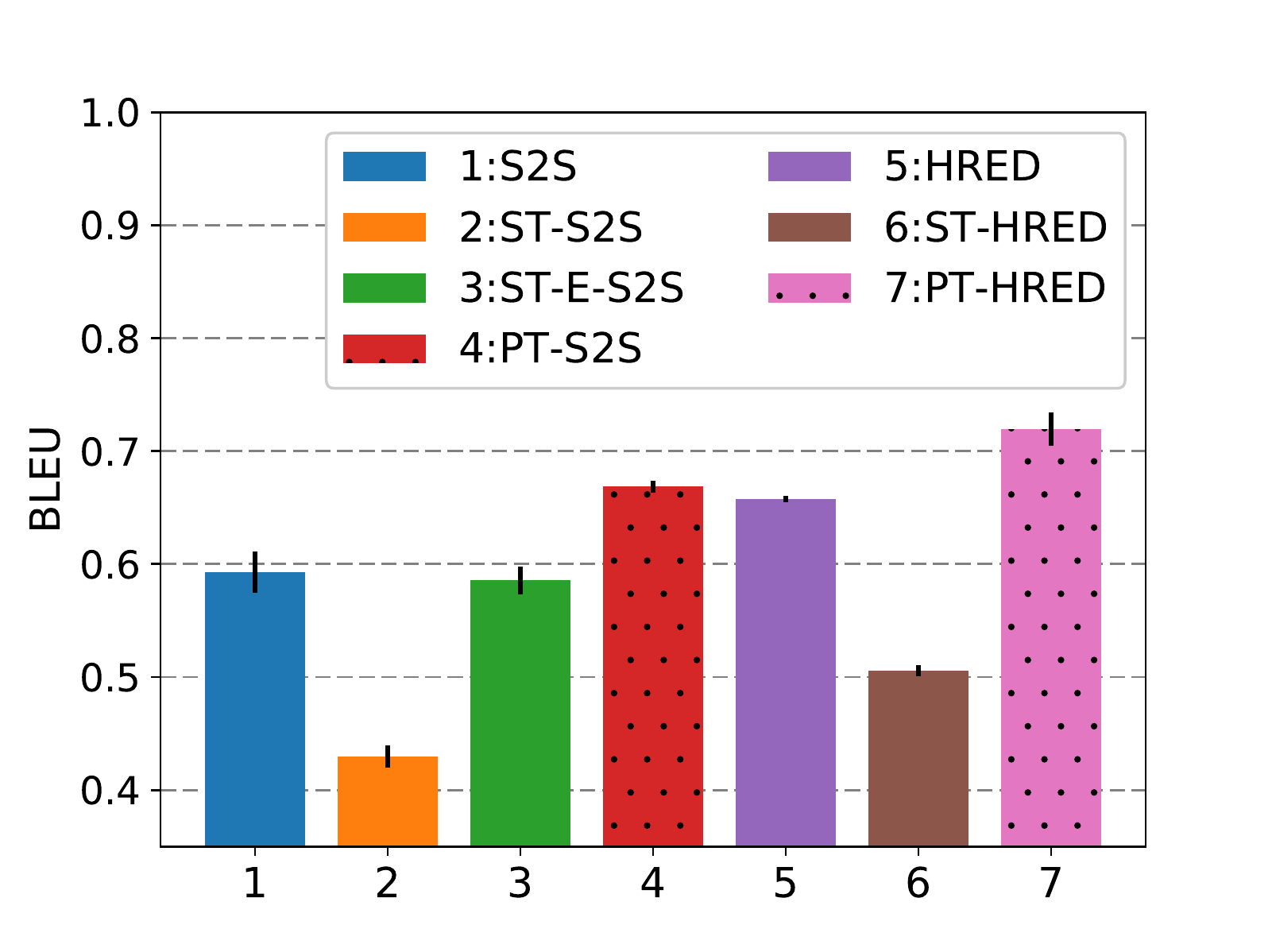}}
}
\caption{BLEU for the simulation dataset}
\label{fig:sim_bleu}
\vspace{-1em}
\end{figure}

\begin{figure}[t]
\centering\mbox{
\scalebox{1.0}{\includegraphics[width=\columnwidth]{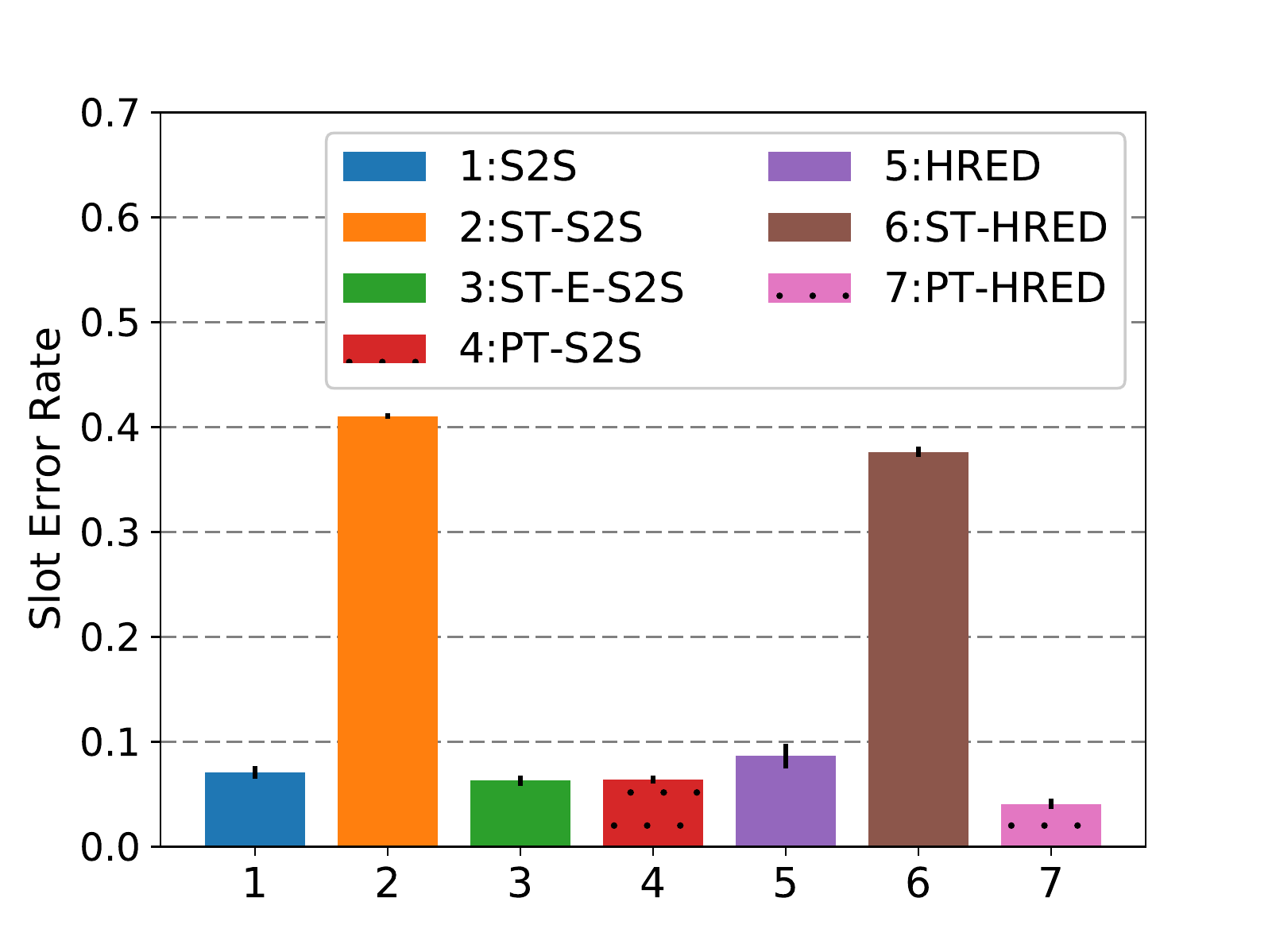}}
}
\caption{Slot-error-rate for the simulation dataset}
\label{fig:sim_slot_error}
\vspace{-1em}
\end{figure}

\begin{figure}[t]
\centering\mbox{
\scalebox{1.0}{\includegraphics[width=\columnwidth]{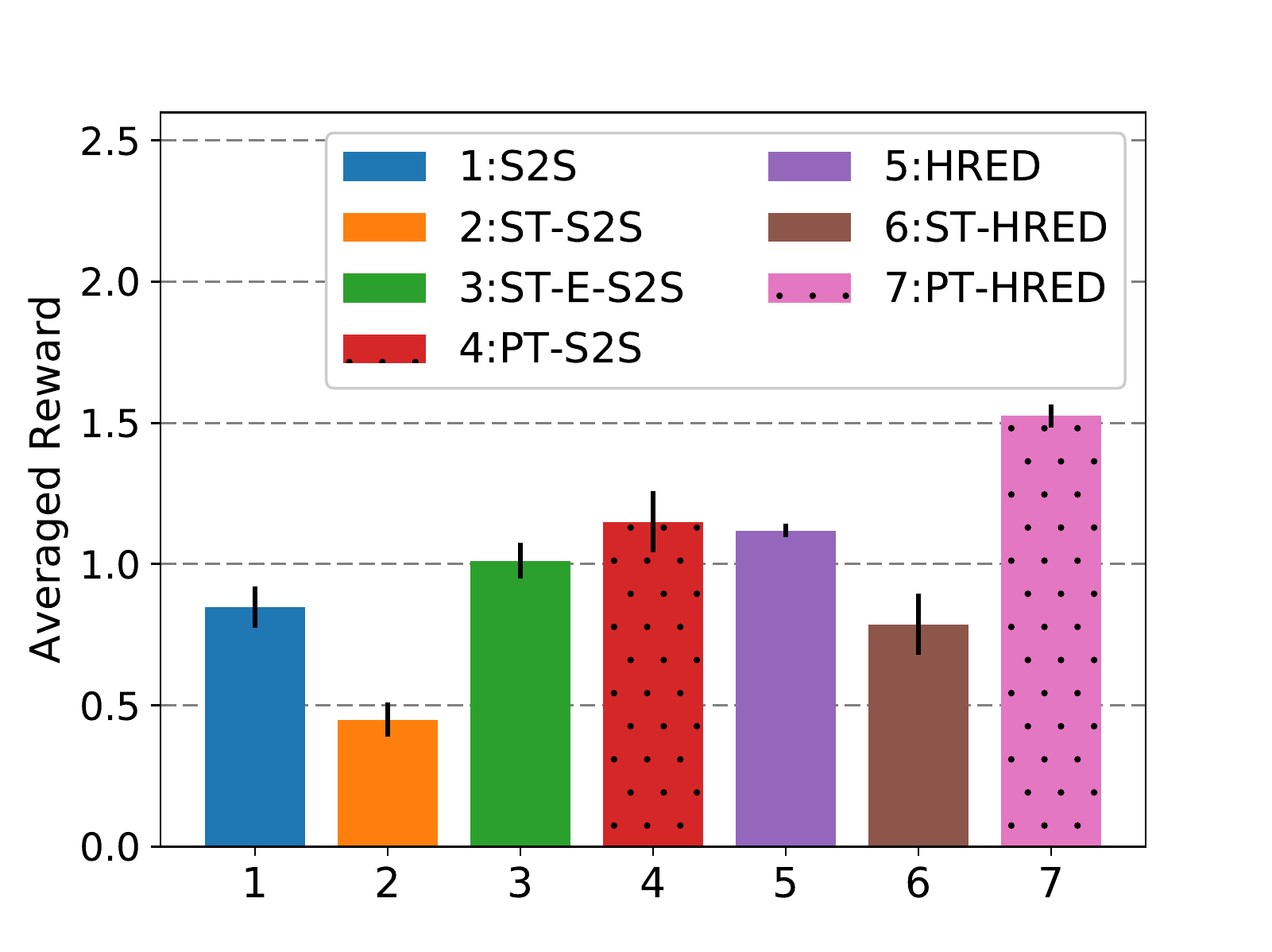}}
}
\caption{Averaged reward for the simulation dataset}
\label{fig:sim_reward}
\vspace{-1em}
\end{figure}

\begin{figure}[t]
\centering\mbox{
\scalebox{1.0}{\includegraphics[width=\columnwidth]{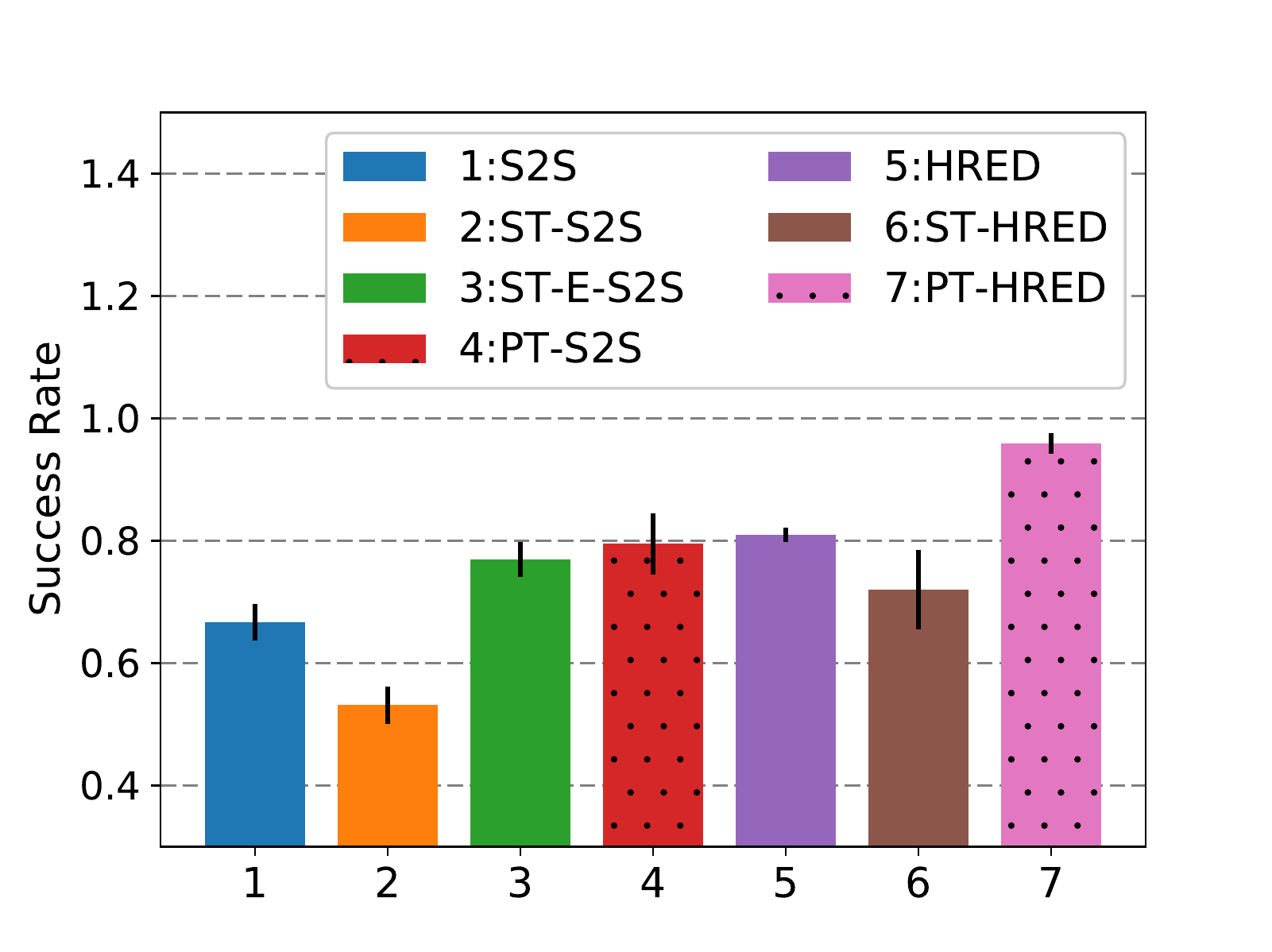}}
}
\caption{Averaged success rate for the simulation dataset}
\label{fig:sim_success}
\vspace{-1em}
\end{figure}

All results are shown in Figures~\ref{fig:sim_bleu}-\ref{fig:sim_success} with details listed in Table~\ref{tab:ResultTab}. The none-transfer models, i.e., ``S2S'' and ``HRED'', work well because the simulated dialogues are relatively simple.
The sentence-level transfer methods ``ST-S2S'', ``ST-E-S2S'' and ``ST-HRED'' perform worst possibly due to the negative transfer. By analyzing the online evaluation records, we found that the sentence-level transfer methods suffer from, for example, using the wrong personal information as shown in the case study section. Personalized phrase-level transfer methods obtain most of the personal information correctly and hence achieve the highest averaged reward, success rate and BLEU scores. These results show that our personalized decoder can improve baseline models and avoid negative transfer caused by domain differences.

\begin{table*}[!ht]
\caption{Experiment Results}
\begin{center}
\begin{small}
\begin{tabular}{|c||c|c|c|c|c|}
  \hline
   & \multicolumn{4}{|c|}{Simulation} & \multicolumn{1}{|c|}{Real data}  \\
  \hline
   & BLEU & Reward & SuccessRate & SlotError & BLEU \\
  \hline
  S2S & 0.5931 $\pm$ 0.0182 & 0.8478 $\pm$ 0.0724 & 0.6667 $\pm$ 0.0302 & 0.0708 $\pm$ 0.0061 & 0.1278 $\pm$ 0.0039  \\
  \hline
  ST-S2S & 0.4297 $\pm$ 0.0099 & 0.4485 $\pm$ 0.0606 & 0.5310 $\pm$ 0.0302 & 0.4103 $\pm$ 0.0026 & 0.1411 $\pm$ 0.0078  \\
  \hline
  ST-E-S2S & 0.5856 $\pm$ 0.0123 & 1.0121 $\pm$ 0.0633 & 0.7691 $\pm$ 0.0285 & $\mathbf{0.0627 \pm 0.0048}$ & 0.1513 $\pm$ 0.0046  \\
  \hline
  PT-S2S & $\mathbf{0.6685 \pm 0.0050}$ & $\mathbf{1.1493 \pm 0.1086}$ & $\mathbf{0.7945 \pm 0.0503}$ & 0.0638 $\pm$ 0.0038 & $\mathbf{0.1609 \pm 0.0055}$  \\
  \hline
  \hline
  HRED & 0.6575 $\pm$ 0.0029 & 1.1187 $\pm$ 0.0241 & 0.8095 $\pm$ 0.0112 & 0.0862 $\pm$ 0.0120 & 0.1387 $\pm$ 0.0032  \\
  \hline
  ST-HRED & 0.5058 $\pm$ 0.0048 & 0.7860 $\pm$ 0.1085 & 0.7199 $\pm$ 0.0650 & 0.3763 $\pm$ 0.0046 & 0.1477 $\pm$ 0.0062  \\
  \hline
  PT-HRED & $\mathbf{0.7193 \pm 0.0148}$  & $\mathbf{1.5247 \pm 0.0413}$ & $\mathbf{0.9587 \pm 0.0164}$ & $\mathbf{0.0405 \pm 0.0051}$ & $\mathbf{0.1588 \pm 0.0093}$ \\
  \hline
\end{tabular}
\label{tab:ResultTab}
\end{small}
\end{center}
\vspace{-1em}
\end{table*}

\begin{figure}[!ht]
\vspace{-1em}
\centering\mbox{
\scalebox{1.0}{\includegraphics[width=\columnwidth]{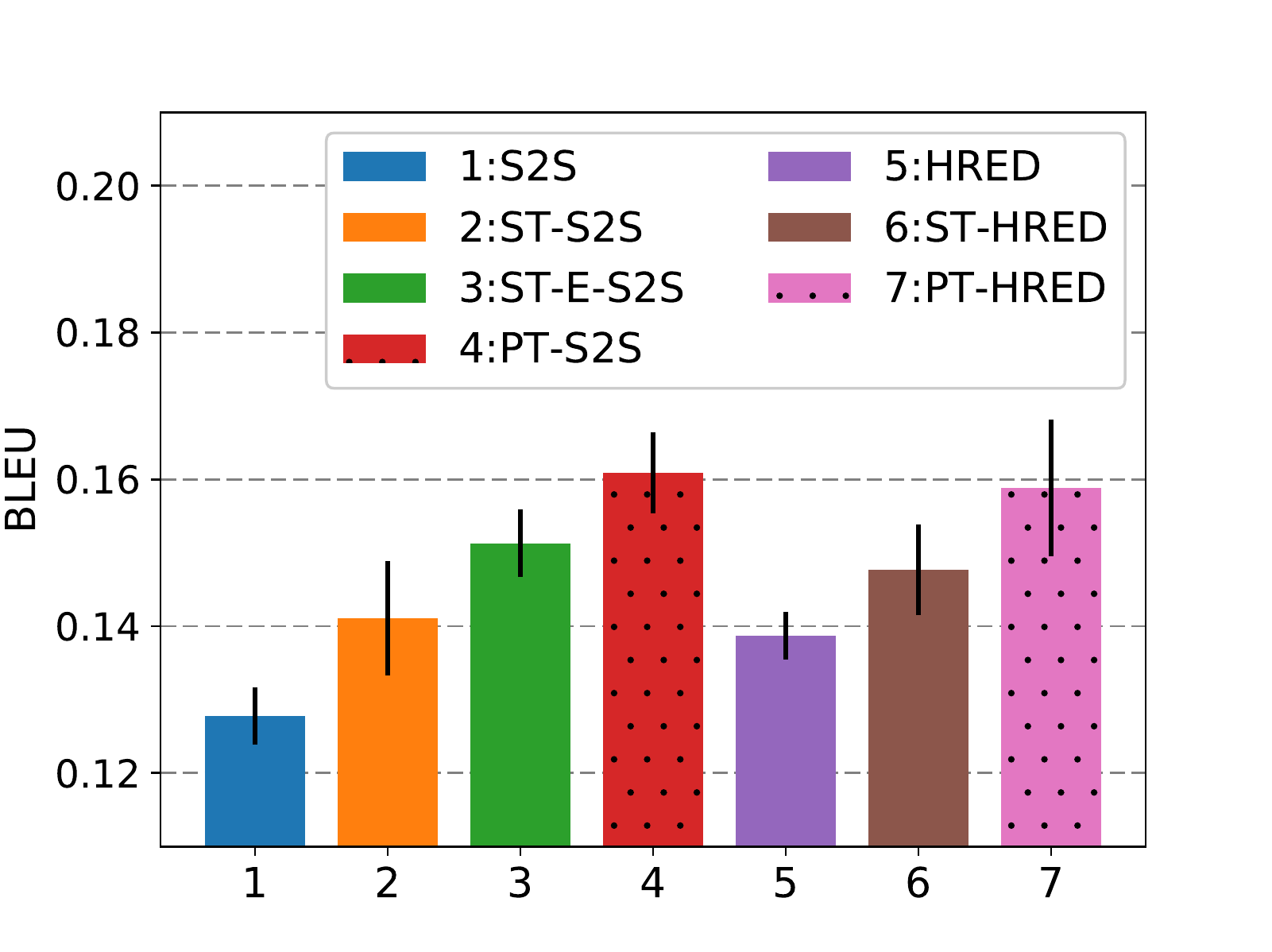}}
}
\caption{BLEU for the real-world dataset}
\vspace{-1em}
\label{fig:real_bleu}
\end{figure}


\subsection{Real-World Experiments}
In this section, we conduct experiments on a real-world dataset.

The dataset is a coffee ordering dataset collected from an O2O service in China. The dialogues are conducted between real customers and real waiters via instant messages. The customers are frequent shoppers, and the waiters are specially trained. On average there are four turns of interactions in each dialogue. In this dataset, there are 52 users with $8.9$ dialogues on average for each user. The statistics of this dataset are listed in Table~\ref{tab:DataTab}.

We also use the BLEU as the evaluation metric. For the BLEU score, 1-gram and 2-gram are used because many sentences are relatively short. Reward and success rate cannot be calculated because policies cannot be run live on static data. 
For each dialogue turn in each testing dialogue, we randomly generate five system responses, and we calculate the averaged BLEU score over the five responses. We train each model with five different random seeds and report the mean and standard deviation of the scores in the test set.

All results are listed in Figure~\ref{fig:real_bleu} with details listed in Table~\ref{tab:ResultTab}.
We can see that ``S2S'' and ``HRED'' have lowest BLEU score, because the training data from each individual user is insufficient for training a competitive model. 
The sentence-level transfer methods ``ST-S2S'', ``ST-E-S2S'' and ``ST-HRED'' have higher BLEU score than none-transfer methods, which demonstrates that transfer learning can indeed help improve the performance.
The personalized phrase-level transfer methods ``PT-HRED'' and ``PT-S2S'' outperform their corresponding sentence-level \mbox{counterparts}, i.e., ``ST-S2S'', ``ST-E-S2S'' and ``ST-HRED'', in terms of BLEU, which demonstrates the effectiveness of the personalized decoder. Specifically, ``PT-HRED'' improves ``ST-HRED'' by $7.5\%$ in terms of BLEU. 
These experimental results again demonstrate that our personalized decoder improves several baseline models and alleviates the negative transfer effect.

\subsection{Case study}
\label{subsect:case}
In this section, we show a case to compare a sentence-level transfer model and a phrase-level transfer model to see how personalized decoder avoids negative transfer. As shown in Table~\ref{tab:DialogCaseTable}, we can see that the sentence-level transfer method ``ST-HRED'' transfers the wrong personal information from other domains and thereby leads to the failure. On the contrary, ``PT-HRED'', the phrase-level transfer method, does transfer the correct personal information. As a result, it successfully avoids negative transfer caused by domain differences.

\begin{table}[ht]
\caption{A case study by comparing the sentence-level transfer model ``ST-HRED'' and the phrase-level transfer model ``PT-HRED'', where the personal words are in bold.}
\begin{center}
\begin{small}
\begin{tabular}{rl}
  \hline
  \multicolumn{2}{c}{\bf{Response Generation Comparison}} \\
  User: & I want a cup of coffee. \\
  Truth Agent: & Same as before, \textbf{tall hot macchiato}\\
  & and deliver to \textbf{Building No.1, Zhengda} \\
  & \textbf{Wudaokou Plaza, No.1199 Minsheng} \\
  & \textbf{Road, Pudong District, Shanghai}? \\
  ST-HRED: & Same as before, tall hot latte \\
  & and deliver to the Fiyta Building, Science \\
  & and Technology Part, Shen Zhen? \\
  PT-HRED: & Same as before, \textbf{tall hot macchiato}\\
  & and deliver to \textbf{Building No.1, Zhengda} \\
  & \textbf{Wudaokou Plaza, No.1199 Minsheng} \\
  & \textbf{Road, Pudong District, Shanghai}? \\
  \hline
\end{tabular}
\label{tab:DialogCaseTable}
\end{small}
\end{center}
\vspace{-2em}
\end{table}

\section{Conclusion}
\label{sect:Conclusion}
Training a personalized dialogue system requires a large amount of data, which is usually unavailable from an individual user. The existing seq2seq transfer learning models operate on the entire sentence, while personalized dialogues require finer granularity.
We propose a personalized decoder, which can transfer phrase-level information between different users while keeping personalized information \mbox{intact}. The proposed personalized decoder can easily be combined with many models including seq2seq and HRED to achieve knowledge transfer. Experimental results demonstrate that the phrase-level transfer personalized models improve the BLEU over multiple sentence-level transfer baseline \mbox{models}. We will study dialogue transfer learning under finer-granularity like prefix and stem in the future.


\bibliography{ref_mo}
\bibliographystyle{aaai}

\end{document}